\documentclass{article}

\usepackage{graphicx}
\usepackage[preprint]{neurips_2023}
\usepackage{fontawesome}
\usepackage{enumitem}

\usepackage[utf8]{inputenc} 
\usepackage[T1]{fontenc}    
\usepackage{hyperref}       
\usepackage{url}            
\usepackage{booktabs}       
\usepackage{amsfonts}       
\usepackage{nicefrac}       
\usepackage{microtype}      
\usepackage{xcolor}         
\usepackage{amsmath} 
\newcommand{\maf}{\text{mafin}}

\title{Mafin: Enhancing Black-Box Embeddings with Model Augmented Fine-Tuning}
\usepackage{natbib}
\setcitestyle{authoryear,round,citesep={;},aysep={,},yysep={;}}
\hypersetup{
    colorlinks=true, 
    urlcolor=blue,    
}

\author{%
  Mingtian Zhang \textsuperscript{1,2}\thanks{Correspondence to: Mingtian Zhang <mingtian@vectify.ai>.}\And
     Shawn Lan \textsuperscript{1} \And
     Peter Hayes\textsuperscript{2} \And
     David Barber \textsuperscript{1,2}
  \Aff
\noindent 
\textsuperscript{1}
Vectify AI \qquad \textsuperscript{2}University College London}

\begin{document}

\maketitle

\begin{abstract}
Retrieval Augmented Generation (RAG) has emerged as an effective solution for mitigating hallucinations in Large Language Models (LLMs). The retrieval stage in RAG typically involves a pre-trained embedding model, which converts queries and passages into vectors to capture their semantics. However, a standard pre-trained embedding model may exhibit sub-optimal performance when applied to specific domain knowledge, necessitating fine-tuning. This paper addresses scenarios where the embeddings are only available from a black-box model. We introduce Model augmented fine-tuning (Mafin) -- a novel approach for fine-tuning a black-box embedding model by augmenting it with a trainable embedding model. Our results demonstrate that Mafin significantly enhances the performance of the black-box embeddings by only requiring the training of a small augmented model. We validate the effectiveness of our method on both labeled and unlabeled datasets, illustrating its broad applicability and efficiency.

\end{abstract}

\section{Introduction}

Large language models (LLMs)~\citep{brown2020language, chowdhery2023palm, achiam2023gpt, team2023gemini, chen2021evaluating} have demonstrated remarkable capabilities across a variety of language tasks. Pre-trained on extensive datasets, these models encapsulate a vast range of world and domain-specific knowledge within their parameters. Despite these advancements, two challenges significantly impact the practical application of pre-trained LLMs:
\begin{enumerate}
    \item They are limited to the knowledge contained within their training datasets. To incorporate up-to-date or private data, fine-tuning the pre-trained LLM is necessary, which can incur a substantial computational cost.
    \item They may produce inaccurate content or ``hallucination'' that significantly diverge from the facts presented in the training data~\citep{bang2023multitask, manakul2023selfcheckgpt, lightman2023let}.
\end{enumerate}

Retrieval-augmented generation (RAG)~\citep{lewis2020retrieval} integrates pre-trained language models with advanced information retrieval techniques, significantly expanding the knowledge base accessible to these models~\citep{gao2023retrieval,jiang2023active,siriwardhana2023improving,gupta2024rag}. This strategy enhances LLMs by utilizing data from external databases and documents~\citep{dinan2018wizard,qin2019conversing,lian2019learning}. A key feature of RAG is its ability to extract knowledge from various corpora, allowing the system to easily integrate new knowledge and effectively address the hallucination issues commonly associated with LLMs~\citep{shuster2021retrieval,wu2023ragtruth,lyu2024crud}.
RAG's proficiency is evident across multiple NLP tasks, including question answering~\citep{cai2018skeleton,cai2019retrieval,su2021prototype}, machine translation~\citep{cai2021neural,xia2019graph,gu2018search}, and code generation~\citep{hashimoto2018retrieve}.

The key component in a RAG system is the information retrieval procedure. Traditionally, researchers have employed lexical methods such as TF-IDF and BM25~\citep{robertson2009probabilistic} for text information retrieval. However, these lexical methods are somewhat rigid, troubled by the lexical gap~\citep{berger2000bridging}, capturing only the similarity in individual or distribution of words, failing to retrieve based on the semantics of the text.  Recently, researchers have introduced dense retrieval methods, mapping queries, and documents to a shared embedding space~\citep{gillick2018end}. These are capable of capturing semantic matches, exhibiting robust performance in various question-answering tasks~\citep{guo2020multireqa,karpukhin2020dense,luan2021sparse}. We give a brief introduction below.

\subsection{Information Retrieval with Text Embedding}

Given an input query $q$, the retriever aims to retrieve a list of relevant passages from a knowledge base $\mathcal{D}=\{x_1,\cdots, x_m\}$. We denote the embedding model as $e(\cdot)$, which can map all the passages to embeddings in the vector space $\mathbb{R}^D$. At retrieval time, the same encoder\footnote{We only consider the setting where the same encoder is used to embed both queries and passages, as this is the most common use case in practical applications. Other dual-encoder architectures exist~\citep{dong2022exploring}, which we do not consider in this study.} is used to map the query $q$ into the same embedding space. The relevance score, which we denote as $s$, between the query and the passages can be calculated using the cosine similarity score in the embedding space
\begin{align}
    s(x,q)=\frac{e(x)^Te(q)}{\lVert e(x)\rVert_2\lVert e(q)\rVert_2},
\end{align}
where $\lVert\cdot \Vert_2$ is the L2 norm. The cosine similarly is equal to the dot product $e^T(x)e(q)$ if the embeddings are normalized, i.e. $\lVert e(x)\rVert_2=1$. 
In practice, the top-K most relevant passages are retrieved, ranked by their relevance scores, containing the relevant knowledge of the query.

Typically, a pre-trained embedding model is employed to convert passages and queries into vector representations. However, these embeddings often less effective when applied to new domains, such as in medical and financial applications~\citep{thakur2021beir,muennighoff2022mteb}. 
 In these cases, it is often advantageous to fine-tune the embedding models on domain-specific data to increase the retrieval accuracy~\citep{gupta2024rag,ovadia2023fine,shi2023replug}. 
 However, some high-performing embedding models, such as OpenAI's text-embedding-ada-002\footnote{See \url{https://platform.openai.com/docs/guides/embeddings} for an introduction.}, have not been made available as open-source. In such cases, users are limited to retrieving embeddings via an API, without access to the model's parameters, making direct fine-tuning of these black-box models unfeasible. To tackle this challenge, this paper presents a novel embedding fine-tuning framework,  named \textbf{M}odel \textbf{a}ugmented \textbf{fin}e-tuning (\textbf{Mafin}). This framework allows for the enhancement of black-box embedding models through domain-specific fine-tuning, which we introduce below.

\section{Model Augmented Fine-Tuning}
We denote the black-box embedding model by $e_{\text{bb}}(\cdot)$, where we have access only to its output embedding; provided by an API. We introduce an auxiliary ``white-box'' trainable model $e_{\theta}(\cdot)$, where we can access its parameters to conduct fine-tuning on new data. 
Specifically, assume the embeddings given by $e_{\text{bb}}$ are normalized, i.e. $\lVert e_{\text{bb}}(\cdot)\rVert_2=1$, we let $e_{\theta}(\cdot)$ be a trainable self-normalized model $e_{\theta}(\cdot)$ and define the new embedding function, denoted as $e_{\text{mafin}}$, as
 \begin{align}
     e_{\text{mafin}}(\cdot)=\text{concat}[ e_{\text{bb}}(\cdot),e_{\theta}(\cdot)]/\sqrt{2}.\label{eq:vanilla:mafine}
 \end{align}
 Then $e_{\text{mafin}}(\cdot)$ will also have norm 1. The constant $\sqrt{2}$ can be ignored in practice since it won't affect the rank of the retrieved items. In this case, the cosine similarity is equivalent to the sum of two independent dot products 
\begin{align}
    e^T_{\maf} (x)e_{\text{mafin}}(y)=\frac{e^T_{\text{bb}}(x)e_{\text{bb}}(y)+e^T_{\theta}(x)e_{\theta}(y)}{2}.\label{eq:equal:mafin}
\end{align}
This design effectively increases the representational capacity of the pre-trained, large-scale black-box model while also benefiting from the domain-specific adaptability afforded by the fine-tuning of $e_{\theta}$.

A more flexible score can be achieved by including a weighted sum in the inner products in Equation~\ref{eq:equal:mafin}. For example, we can define the $e_{\lambda\maf}$ embedding with the dot product
\begin{align}e^T_{\lambda\text{-}\maf} (x)e_{\lambda\text{-mafin}}(y)
=\lambda_1(x,y)e^T_{\text{bb}}(x)e_{\text{bb}}(y)+\lambda_2(x,y)e^T_{\theta}(x)e_{\theta}(y),\label{eq:weighted:score}
\end{align}
where $\lambda_1(x,y),\lambda_2(x,y)$  are scalar-valued weights that depend on the input $x,y$. To avoid introducing an additional weighting network to learn $\lambda$, we can use an unnormalized embedding representation, which we denote as $\hat{e}_\theta(\cdot)$ and define $e_{\lambda\maf}$ as 
\begin{align}
e_{\lambda\text{-}\maf}=\text{concat}[ e_{\text{bb}}(\cdot),\hat{e}_{\theta}(\cdot)]/\sqrt{1+\lVert \hat{e}_{\theta}(x)\rVert_2^2}\label{eq:lambda:mafine}
\end{align}
In this case, the normalized embedding can be written as $e_{\theta}(x)=\hat{e}_\theta(x)/\lVert \hat{e}_{\theta}(x)\rVert_2$, which allows us to derive the corresponding analytical form of the $\lambda$s in Equation~\ref{eq:weighted:score}:
\begin{equation}
\small
    \lambda_1(x,y)=\frac{1}{\sqrt{1+\lVert \hat{e}_{\theta}(x)\rVert_2^2}\sqrt{1+\lVert \hat{e}_{\theta}(y)\rVert_2^2}},\quad \lambda_2(x,y)=\frac{\lVert \hat{e}_{\theta}(x)\rVert_2\lVert \hat{e}_{\theta}(y)\rVert_2}{\sqrt{1+\lVert \hat{e}_{\theta}(x)\rVert_2^2}\sqrt{1+\lVert \hat{e}_{\theta}(y)\rVert_2^2}}.
\end{equation}
Intuitively, $e_{\lambda\maf}$ use the norm of the vector as the additional degree of freedom, 
which allows us to implicitly learn both $\lambda_1(x,y)$ and $\lambda_2(x,y)$ without introducing additional networks. 

The training criteria for learning the $e_\theta$ in the Mafin model are tailored to the specific requirements of the task depending on whether we can access the relevance labels between the query and the candidate passages.

\section{Fine-Tuning Criteria\label{sec:drl}}
Fine-tuning the retrieval model with relevance labels is akin to solving a learning-to-rank problem~\citep{herbrich1999support, freund2003efficient, burges2005learning}. The field of information retrieval has introduced numerous methods for learning to rank, with a comprehensive overview provided by~\cite{liu2009learning}. In our study, we concentrate on a probabilistic framework previously discussed by~\cite{cao2007learning,xia2008listwise} and we introduce an efficient approximation method that supports training with large-scale corpora. Furthermore, we explore scenarios where queries and relevance labels are unavailable. In response to this challenge, we propose utilizing a pre-trained LLM to generate question-passage pairs, which enables the application of learning-to-rank techniques by providing necessary relevance for model training.

\subsection{Probabilistic Ranking}
We assume for each query $q$ and a candidate passage $x_k$, we can access the integer relevance label $y_k$, e.g. $y_k\in \{0,1,\cdots\}$, where a higher value indicates a more relevant passage. Given a query $q$, a list of passages $[x_1,x_2,\cdots,x_K]$ with a relevant label list $[y_1,y_2,\cdots,y_K]$, we can define a mapping $\pi$ that maps from the canonical index list $[1,\cdots, K]$ to a ranked index list  $[\pi(1),\cdots, \pi(K)]$ based on the relevance score from high to low. Therefore, we can create a surrogate ground truth distribution of the permutation, which we denote as $p_s(\pi|q)$, represented by
 \begin{align}
p_s(\pi|q)=\prod_{k'=1}^K\frac{\exp\left(y_{\pi^{-1}(k')}/\tau)\right)}{
    \sum_{u=k'}^K \exp(y_{\pi^{-1}(u)}/\tau)
    },
\end{align}
where we use $k'$ to denote the rank in the ranked list $k'\in [\pi(1),\cdots, \pi(K)]$  and where $\tau$ is the temperature. We can also build a model based on the embedding model $e_\theta(\cdot)$ and relevance score $s_\theta(\cdot, \cdot)=e^T_\theta(\cdot)e_\theta(\cdot)$ in a similar fashion 
 \begin{align}
p_\theta(\pi|q)=\prod_{k'=1}^K\frac{\exp\left(s_\theta(q, x_{\pi^{-1}(k')})\right)}{
    \sum_{u=k'}^K \exp(s_\theta\left(q, x_{\pi^{-1}(u)})\right)
    }
\end{align}
This model is also referred to as the Plackett-Luce model~\citep{plackett1975analysis,luce2005individual}. 
To learn the parameter $\theta$, we can minimize the KL divergence
\begin{align}
    \mathrm{KL} (p_s(\pi|q)|| p_\theta(\pi|q)) = -\sum_{\Pi} p_s(\pi|q)\log p_\theta(\pi|q) + const., \label{eq:mle}
\end{align}
where $\Pi$ denotes the set of all possible permutations. However, for a large $K$, explicitly summing over $\Pi$ will be intractable. The paper~\citep{cao2007learning} proposes to use a top-1 approximation by only considering the probability of some chunk being ranked in the top-1 position, which leads to a top-1 surrogate ground truth distribution and a top-1 ranking model 
 \begin{equation}
 \small
p_s^{\text{Top}1}(\pi(k)=1|q)=\frac{\exp\left(y_{k}/\tau)\right)}{
    \sum_{u=1}^K \exp(y_{k}/\tau)
    },\quad p^{\text{Top}1}_\theta(\pi(k)=1|q)= \frac{\exp\left(s_\theta(q, x_k)\right)}{
    \sum_{u=1}^K \exp(s_\theta\left(q, x_u)\right)}.\label{eq:top1}
\end{equation}
Then minimizing the KL divergence between  $p_s^{\text{Top}1}(\pi(k)=1|q)$ and $p^{\text{Top}1}_\theta(\pi(k)=1|q)$ can be simplified to 
\begin{align}
    \mathrm{KL} (p^{\text{Top}1}_s|| p^{\text{Top}1}_\theta) = -\sum_{k=1}^K p^{\text{Top}1}_s(\pi(k)=1|q)\log p^{\text{Top}1}_\theta(\pi(k)=1|q)+const.,
\end{align}
However, for large passage number $K$, summing over $K$ class and calculating the normalization over $K$ classes can still cause a computation bottleneck. We therefore now propose an efficient approximation method to Equation~\ref{eq:top1}.

\subsubsection{Efficient Approximation For  Large $K$}
 We notice that in most of the datasets used in practice, the relevance score list is sparse, i.e. most of the relevance scores of the candidate passages for the given query is $0$. This phenomenon allows us  
to further improve the computation efficiency by letting $\tau\rightarrow 0$, so $p_d(\pi(k)=1|q)$ becomes a delta distribution. Consequently, we have
\begin{align}
    -\sum_{k=1}^M p^{\text{Top}1}_s(\pi(k)=1|q)\log p^{\text{Top}1}_\theta(\pi(k)=1|q)\xrightarrow{ \tau\rightarrow 0} -\log p^{\text{Top}1}_\theta(\pi(k_{\text{max}})=1|q),\label{eq:top1:mle}
\end{align}
where 
\begin{align}
k_{\text{max}}={\arg\max}_{k\in \{1,\cdots,K\}} \log p^{\text{Top}1}_s(\pi(k)=1)={\arg\max}_{k\in [1,\cdots, K]} y_k.
\end{align}
In practice, when $K$ is very large, we can also sample a mini-batch with a size of $M\ll K$ and use the sampled sub-list to replace the full-list used in Equation~\ref{eq:top1:mle}, which we found works well in practice.
Intuitively, this process aims to adjust the model so that relevant query-passage pairs are associated with higher similarity scores and irrelevant pairs with lower scores. Surprisingly, we find this training objective is also equivalent to the InfoNCE loss~\citep{oord2018representation} in the contrastive learning literature, which has been successfully used to pre-train  embedding models~\citep{lee2019latent,izacard2021unsupervised}. Therefore, our formulation provides a probabilistic justification for the relationship between the learning-to-rank loss and the contrastive learning loss, a connection not previously discussed to our knowledge.

\subsection{Fine-Tuning without Labels}
In scenarios where only candidate passages are accessible, without any training queries or corresponding relevance labels, the learning-to-rank approach discussed previously becomes inapplicable. To overcome this challenge, we utilize the capabilities of large language models (LLMs), such as ChatGPT, to generate synthetic queries for fine-tuning embeddings. Specifically, for a given passage ${x_1, \cdots, x_N}$, we generate a corresponding set of queries ${q_1, \cdots, q_N}$ using the following prompt:
\begin{align}
q_n \sim p_{\text{LLM}}(q| \text{prompt}=\text{`generate a query based on the given passage'}, x_n).
\end{align}
After generating these query-passage pairs ${(q_1, x_1), \cdots, (q_N, x_N)}$, we proceed under the assumption that each query $q_n$ is relevant to its corresponding passage $x_n$, but not to any other passages $x_{i \neq n}$ in the set. This approach is a specialized case of the scenarios discussed previously, employing binary relevance labels $y \in {0, 1}$, and transforming the relevance label list $[y_1, \cdots, y_N]$ into a one-hot list for each query $q_n$. This leads to a simplified form of Equation~\ref{eq:top1:mle}:
\begin{align}
p_\theta(y_k=1 | x_k, q) = \frac{\exp(s_\theta(x_k, q))}{\sum_{k=1}^K \exp(s_\theta(x_k, q))}. \label{eq:mle}
\end{align}
In this setup, the fine-tuning task evolves into a single-label classification task, enabling us to conduct maximum likelihood learning (MLE) by maximizing the log of Equation~\ref{eq:mle} on the training dataset to fine-tune the embedding model.

\section{Empirical Demonstrations}
In this section, we empirically investigate the performance of the proposed Mafin method.
\subsection{Dataset}

We employ two widely-used text retrieval datasets, FiQA-2018~\citep{maia201818} and NFCorpus~\citep{boteva2016full}, to assess the effectiveness of our approach. The detailed descriptions of these two datasets are as follows:
\setlist[itemize]{ nolistsep}
\begin{itemize}
\item \textbf{FiQA-2018:}
    The FiQA-2018 dataset is specialized for the financial domain, featuring question-and-answer pairs derived from Stackexchange posts on investment topics, spanning 2009 to 2017. It includes 57,640 answer posts and is segmented into 17,110 question-answer pairs for training, plus 531 pairs for testing. Each question is associated with an average of 2.6 relevant posts. Relevance is binary, rated as either 0 or 1, indicating whether an answer is relevant to its corresponding question.

\vspace{0.2cm}

\item \textbf{NFCorpus:}
The NFCorpus dataset is dedicated to medical information retrieval, collected from the NutritionFacts.org website. Queries are formulated in everyday language, whereas the corpus comprises scientific articles. The final dataset contains 3,672 answer posts, with relevance levels between the corpus and queries categorized into three levels: 0, 1, and 2. On average, each query is associated with 38.2 related posts.

\end{itemize}
We follow BEIR~\citep{thakur2021beir} to partition and utilize these two datasets. BEIR divides these datasets into three sets: train, development, and test. On the FiQA dataset, we use the train set for training, the development set for validation, and the test set for testing. On the NFCorpus dataset,  we employ 80\% of the development set, as partitioned by MTEB, for training, 20\% for validation, and the test set for testing.

\subsection{Metrics}
For evaluation, we use the Recall~\citep{buckland1994relationship} and the NDCG~\citep{valizadegan2009learning} metrics throughout all the experiments, which we briefly introduce below.

\textbf{Recall@$K$}: Recall is defined as the ratio of the number of relevant passages retrieved to the total number of relevant passages in the database. It is given by the formula:
    \begin{equation}
    \text{Recall}@K = \frac{\text{Number of Relevant Passages Retrieved in first K items}}{\text{Total Number of Relevant Passages}}
    \end{equation} 
    
    This metric primarily focuses on the completeness of the retrieval process. A high recall value indicates that the system is able to retrieve most of the relevant passages.
    
\textbf{\text{NDCG}@$K$}: NDCG is a measure of ranking quality. It evaluates the effectiveness of a text retrieval system based on the graded relevance of the passages in the result list. The NDCG value ranges from 0 to 1, where 1 represents the ideal ranking of passages. The formula for NDCG at a particular rank position $K$ is: 
    \begin{equation}
    \text{NDCG}@K = \text{DCG}@K/\text{IDCG}@K
    \end{equation}
    where $\text{DCG}@K$ (Discounted Cumulative Gain at rank $K$) is calculated as:
\begin{equation}
\text{DCG} @ K=\sum_{i=1}^K (2^{r e l_i}-1)/\log _2(i+1)
\end{equation}
and $\text{IDCG}@K$ is the Ideal Discounted Cumulative Gain at rank $K$. In $\text{DCG}@K$, $rel_i$ is the graded relevance of the result at position $i$. 
NDCG is particularly useful in scenarios where passages have varying levels of relevance. It is a more comprehensive metric compared to Recall, as it takes into account the position of relevant passages in the search results and the gradation of their relevance.

On the FiQA-2018 dataset, each query corresponds to an average of 2.6 ground truth corpora. We selected values of $K$ as 1, 3, and 5 for this dataset. Conversely, on the NFCorpus dataset, each query corresponds to an average of 38.2 ground truth corpora, significantly more than in the FiQA-2018. Therefore, we chose values of $K$ as 5, 10, and 20 for the NFCorpus dataset.

\subsection{Implementation Details}
 We employ OpenAI's text-embedding-ada-002~\footnote{https://platform.openai.com/docs/guides/embeddings} as our black-box embedding model. For the enhanced models, we select models that exhibit performance gaps but are relatively close to text-embedding-ada-002 on their respective datasets. Specifically, for FiQA-2018, we opt for BAAI's bge-base-en-v1.5 model~\footnote{https://huggingface.co/BAAI/bge-base-en-v1.5}, and for NFCorpus, we choose BAAI's bge-small-en-v1.5 model~\footnote{https://huggingface.co/BAAI/bge-small-en-v1.5}.
The BGE model class offers three scales (large, base, small) of BERT-like~\citep{devlin2018bert} architectures for diverse applications, with the [CLS] token embedding utilized for representation. The models are pre-trained on large-scale datasets using an MAE-style method for foundational training. This is followed by label-free contrastive learning to improve discrimination between text pairs and labeled multi-task learning for fine-tuning toward specific tasks. This structured training approach, leveraging extensive text data, contrastive techniques, and task-oriented optimization, is designed to generate embeddings with wide-ranging utility and capacity for enhanced performance in targeted applications~\citep{xiao2023c}. 

Our experiments are conducted on an NVIDIA 4090 GPU. We implement our algorithm and each baseline using Pytorch~\citep{paszke2019pytorch}. For our method and all baselines that require additional model training, we conduct thorough hyperparameter searches to reveal the optimal performance of our approach. Training is performed on the training dataset, and we decide when to stop based on performance on the validation set. To enhance model generalization and mitigate overfitting, we employed label smoothing in our approach. The label smoothing rate is considered a hyperparameter, and we search the optimal value within the set $\{0, 0.1, 0.2, 0.3, 0.4, 0.5\}$. We employ the early stopping method, which terminates training when there is no improvement in performance on the validation set for four consecutive epochs compared to the previous best-performing epoch.

\subsection{Comparison Methods}
To demonstrate the effectiveness of Mafin, we conduct a comprehensive comparison against several relevant techniques, which we discuss below.

\textbf{Original Black-box Model:} As a baseline, we utilize OpenAI's text-embedding-ada-002 as the black-box model for both the FiQA-2018 and NFCorpus datasets. 

\textbf{Original Augmenting Model (BBAI):} Our approach targets scenarios where the highest-performing embedding model operates as a black box, and where traditional, adjustable embedding models do not meet performance expectations. For our experiments with the FiQA-2018 and NFCorpus datasets, we selected BAAI's bge-base and bge-small, respectively, as the augmenting models. 

\textbf{Only Fine-tuning:} This method involves solely fine-tuning the augmenting model with the training dataset, aiming to enhance its performance without leveraging the Original Black-box Model.

\textbf{Simple Concatenation:} We assess the effectiveness of a straightforward strategy that involves concatenating the embeddings from the Black-box Model with those from the Augmenting Model, without any additional modifications or fine-tuning.

\textbf{Concatenation after Fine-tuning:} This method enhances the simple Concatenation approach by first fine-tuning the augmenting model on the training dataset, and then concatenating its output with the output of the Black-box Model.

\textbf{Linear Matrix Fine-tuning.}
An alternative way to fine-tune the black-box embedding model~\footnote{https://github.com/openai/openai-cookbook/blob/main/examples/Customizing\_embeddings.ipynb} is to add a linear transformation on the black-box embedding 
$$e_{\text{new}}(\cdot)=W e_{\text{bb}}(\cdot),$$
where $W\in\mathbb{R}^{D\times D}$ is a learnable linear matrix. For scenarios where $D$ is large, we consider using a low-rank approximation $W = W_l W_r^T$, where $W_l, W_r \in \mathbb{R}^{D {\times} R}$ and $R < D$. This method, however, has limitations, such as the potential for information loss when $W$ is non-invertible and the lack of input-dependence in the transformation matrix $W$. To address domain-specific adaptation, we initialize a transformation matrix, aiming to map the black-box model's embeddings into a domain-specific space. This matrix, of dimension $d {\times} d$, is initially set close to an identity matrix to preserve the original embedding information while allowing for learnable adjustments.

The effectiveness and limitations of these methods are further discussed in the results section, providing a detailed comparison of their performance across different datasets.

\subsection{Experiment Results of Mafin}

\begin{table}[]
\center
\small
\caption{Performance of Mafin and comparison methods on FiQA-2018 dataset. 
Performance Gain refers to the percentage increase in performance of our method compared to only fine-tuning.}
\begin{tabular}{ccccccc}
\hline
\textbf{Model}                 & \textbf{Recall@1} & \textbf{Recall@3} & \textbf{Recall@5} & \textbf{NDCG@1} & \textbf{NDCG@3} & \textbf{NDCG@5} \\ \hline
Black Box Model                & 0.38362           & 0.60531           & 0.68946           & 0.70216         & 0.77010         & 0.77973         \\
Augmenting Model      & 0.35477           & 0.58500           & 0.65625           & 0.68364         & 0.75454         & 0.76086         \\
Only Finetune               & 0.38179           & 0.60733           & 0.69165           & 0.70988         & 0.78345         & 0.79359         \\
Direct concat.          & 0.39475           & 0.62353           & 0.70620           & 0.72377         & 0.79268         & 0.79991         \\
Finetune \& concat. & 0.39969           & 0.63864           & 0.71640           & 0.73303         & 0.80420         & 0.81175         \\
Linear Transform.         & 0.37680           & 0.61205           & 0.69180           & 0.68673         & 0.76579         & 0.77368         \\ \hline
Mafin/Unsup.            & 0.39743           & 0.62731           & 0.71310           & 0.72531         & 0.80137         & 0.80877         \\
Improvement & +3.65\% &3.63\%&+3.43\%	&+3.30\%	&+4.06\% &+3.72\% \\ \hline
Mafin/Sup.               & 0.40584           & 0.64684           & 0.72460           & 0.74228         & 0.81344         & 0.82018\\
Improvement &
+5.79\%&+6.86\%&+5.10\%&5.71\%&+5.63\%&+5.19\%\\\hline
\end{tabular}

\label{tab:results_fi}
\end{table}

\begin{table}[]
\center
\small
\caption{Performance of Mafin and comparison methods on NFCorpus dataset. 
Performance Gain refers to the percentage increase in performance of our method compared to only fine-tuning.}
\begin{tabular}{ccccccc}
\hline
\textbf{Model}                       & \textbf{Recall@5} & \textbf{Recall@10} & \textbf{Recall@20} & \textbf{NDCG@5} & \textbf{NDCG@10} & \textbf{NDCG@20} \\ \hline
Black-box model   & 0.14107 & 0.18520 & 0.22506 & 0.55143 & 0.54844 & 0.53401 \\
Augmenting model& 0.13026           & 0.16372            & 0.20085            & 0.53164         & 0.51591          & 0.50836          \\
Only finetune                & 0.13437 & 0.17533 & 0.21825 & 0.52779 & 0.52442 & 0.51025 \\
Direct concat.           & 0.13897 & 0.18563 & 0.22860 & 0.55458 & 0.54989 & 0.53340 \\
Finetune \& concat. & 0.14264 & 0.18370 & 0.23039 & 0.54699 & 0.54357 & 0.52705 \\
Linear Transform.          & 0.14450 & 0.18532 & 0.22723 & 0.50245 & 0.54947 & 0.53353 \\ \hline
Mafin/Unsup.             & 0.13962 & 0.18662 & 0.22868 & 0.55650 & 0.55168 & 0.53727 \\
Improvement  & -1.03\% & +0.77\% & +1.61\% & +0.92\% & +0.57\% & +0.61\% \\ \hline
Mafin/Sup.               & 0.14273 & 0.18956 & 0.23109 & 0.55896 & 0.55355 & 0.53870 \\
Improvement    & +1.18\% & +2.35\% & +2.68\% & +1.57\% & +0.93\% & +0.88\%  \\ \hline
\end{tabular}

\label{tab:results_bio}
\end{table}
We first conduct a comprehensive set of experiments with the vanilla Mafine embedding described in Equation~\ref{eq:vanilla:mafine}. The experimental results are presented in Table~\ref{tab:results_fi} and Table~\ref{tab:results_bio}. The performance gains displayed in the table represent the performance improvements achieved by our method compared to fine-tuning the augmenting model alone.

Through our experiments, we've observed significant advancements in performance with our method over the straightforward strategy of merely fine-tuning an augmenting model for tasks requiring recall in domains where resources are scarce. Fine-tuning open-source models, while simple and direct, significantly underperforms compared to our approach. Our approach consistently delivers at least a 3\% average improvement across all measured metrics in both supervised and unsupervised settings.

Furthermore, when supervised training is possible due to the availability of query-corpus ground truth pairs, our method significantly boosts performance beyond that of the original, less flexible black-box model. By integrating and training the transparent parts of an open-source model alongside the black-box model, we've seen stable enhancements across various metrics on two distinct datasets, surpassing the black-box model's original performance.

In the realm of unsupervised learning, where such ground truth pairs are absent, the proposed approach still achieves notable gains over the original black-box model. By generating question-answer pairs from text segments using large language models and using these for training, we've consistently observed performance improvements, underscoring the unsupervised method's efficacy.

Interestingly, a simple strategy of concatenating the black-box model with the augmenting model also leads to performance enhancements. Despite the augmenting model's inferior performance compared to the black-box model on all metrics within the FiQA-2018 dataset, this concatenation strategy allows it to surpass the original black-box model's performance. This outcome is replicated in the NFCorpus dataset, where the concatenated model generally outperforms the black-box model, suggesting that this approach may leverage a principle similar to bagging in traditional machine learning, where combining multiple different weak classifiers creates a stronger classifier.

Lastly, our investigation into the utility of linear transformations has shown that they do not consistently offer performance improvements. While they may sometimes exceed the original black-box model's performance on certain metrics, they often lead to a decline in performance across the majority of metrics. This inconsistency could be due to the inherent limitations and rigidity of linear transformation methods, limiting their effectiveness in enhancing model performance.

\subsection{Comparison with $\lambda$-Mafine}
We further assess the performance of the standard Mafin embedding, as outlined in Equation~\ref{eq:vanilla:mafine}, against the $\lambda$-mafin embedding, detailed in Equation~\ref{eq:lambda:mafine}. In our $\lambda$-mafin experiment, we select the initial values of $\lambda$s so that the norms of the embeddings from both the Black-box Model and the Augmenting Model are equal.
The comparative outcomes for both the FiQA-2018 and NFCorpus datasets, utilizing these respective strategies, are presented in Table~\ref{tab:lambda_fi} and Table~\ref{tab:lambda_bio}. 
Although the $\lambda$-mafin, in principle, offers increased flexibility through the introduction of additional weighting parameters, experimental findings suggest that its performance is comparable to that of the standard Mafin when both embedding models are assigned equal weights. Therefore, the exploration of the potential advantages offered by this additional parameter is deferred to future research.
\begin{table}[]
\center
\small
\caption{Performance of our model when $\lambda$ is set to 1 and when $\lambda$ is trainable on FiQA-2018 dataset.}
\begin{tabular}{ccccccc}
\hline
\textbf{Model}                 & \textbf{Recall@1} & \textbf{Recall@3} & \textbf{Recall@5} & \textbf{NDCG@1} & \textbf{NDCG@3} & \textbf{NDCG@5} \\ \hline
$\lambda$-mafin, Unsup. & 0.38825           & 0.62921           & 0.71441           & 0.71605         &  0.79956         & 0.80568         \\
Mafin, Unsup.          & 0.39743           & 0.62731           & 0.71310           & 0.72531         & 0.80137         & 0.80877         \\ \hline
$\lambda$-mafin, Sup.            &  0.40527           & 0.63605           & 0.72878           & 0.74074         & 0.80584         & 0.81664         \\
Mafin, Sup.              & 0.40584           & 0.64684           & 0.72460           & 0.74228         & 0.81344         & 0.82018\\ \hline
\end{tabular}

\label{tab:lambda_fi}
\end{table}

\begin{table}[]
\center
\small
\caption{Performance of our model when $\lambda$ is set to 1 and when $\lambda$ is trainable on NFCorpus dataset.}
\begin{tabular}{ccccccc}
\hline
\textbf{Model}                 & \textbf{Recall@1} & \textbf{Recall@3} & \textbf{Recall@5} & \textbf{NDCG@1} & \textbf{NDCG@3} & \textbf{NDCG@5} \\ \hline
$\lambda$-mafin, Unsup. & 0.13625           & 0.18747           & 0.22958           & 0.55705         & 0.55203         & 0.53806         \\
Mafin, Unsup.      &  0.13962 & 0.18662 & 0.22868 & 0.55650 & 0.55168 & 0.53727         \\ \hline
$\lambda$-mafin, Sup.  &  0.14133  & 0.19290 & 0.23221 & 0.55822 & 0.55421 & 0.53960         \\
Mafin, Sup.              & 0.14273 & 0.18956 & 0.23109 & 0.55896 & 0.55355 & 0.53870\\ \hline
\end{tabular}

\label{tab:lambda_bio}
\end{table}

\section{Conclusion}
In this work, we have introduced Mafin, a novel methodology for fine-tuning black-box embedding models, thereby addressing a significant gap in the field of Retrieval Augmented Generation (RAG). Recognizing the need for enhanced performance in black-box embedding models, especially when applied to new documents or within specific domains, Mafin effectively meets this challenge by augmenting a black-box model with a small tunable embedding model thus significantly boosting its performance while only requiring a minimal fine-tuning cost. This method leverages both the powerful language representation provided by large pre-trained models and the benefits of fine-tuning with a small embedding model. The low fine-tuning cost of the small models enables its use for large-scale, customized online fine-tuning tailored to each company and individual, promising to be a performance-effective and cost-efficient framework for the RAG infrastructure.

We have demonstrated Mafin's excellence in fine-tuning embedding models for text retrieval tasks within the RAG framework. Future work will explore Mafin's potential across a broader range of fields. We aim to test and validate our fine-tuning methodology's effectiveness in tasks such as classification and clustering, thereby further expanding the applicability and impact of our approach.
\newpage

\bibliography{ref}

\begin{thebibliography}{55}
\providecommand{\natexlab}[1]{#1}
\providecommand{\url}[1]{\texttt{#1}}
\expandafter\ifx\csname urlstyle\endcsname\relax
  \providecommand{\doi}[1]{doi: #1}\else
  \providecommand{\doi}{doi: \begingroup \urlstyle{rm}\Url}\fi

\bibitem[Achiam et~al.(2023)Achiam, Adler, Agarwal, Ahmad, Akkaya, Aleman, Almeida, Altenschmidt, Altman, Anadkat, et~al.]{achiam2023gpt}
Achiam, J., Adler, S., Agarwal, S., Ahmad, L., Akkaya, I., Aleman, F.~L., Almeida, D., Altenschmidt, J., Altman, S., Anadkat, S., et~al.
\newblock Gpt-4 technical report.
\newblock \emph{arXiv preprint arXiv:2303.08774}, 2023.

\bibitem[Bang et~al.(2023)Bang, Cahyawijaya, Lee, Dai, Su, Wilie, Lovenia, Ji, Yu, Chung, et~al.]{bang2023multitask}
Bang, Y., Cahyawijaya, S., Lee, N., Dai, W., Su, D., Wilie, B., Lovenia, H., Ji, Z., Yu, T., Chung, W., et~al.
\newblock A multitask, multilingual, multimodal evaluation of chatgpt on reasoning, hallucination, and interactivity.
\newblock \emph{arXiv preprint arXiv:2302.04023}, 2023.

\bibitem[Berger et~al.(2000)Berger, Caruana, Cohn, Freitag, and Mittal]{berger2000bridging}
Berger, A., Caruana, R., Cohn, D., Freitag, D., and Mittal, V.
\newblock Bridging the lexical chasm: statistical approaches to answer-finding.
\newblock In \emph{Proceedings of the 23rd annual international ACM SIGIR conference on Research and development in information retrieval}, pp.\  192--199, 2000.

\bibitem[Boteva et~al.(2016)Boteva, Gholipour, Sokolov, and Riezler]{boteva2016full}
Boteva, V., Gholipour, D., Sokolov, A., and Riezler, S.
\newblock A full-text learning to rank dataset for medical information retrieval.
\newblock In \emph{Advances in Information Retrieval: 38th European Conference on IR Research, ECIR 2016, Padua, Italy, March 20--23, 2016. Proceedings 38}, pp.\  716--722. Springer, 2016.

\bibitem[Brown et~al.(2020)Brown, Mann, Ryder, Subbiah, Kaplan, Dhariwal, Neelakantan, Shyam, Sastry, Askell, et~al.]{brown2020language}
Brown, T., Mann, B., Ryder, N., Subbiah, M., Kaplan, J.~D., Dhariwal, P., Neelakantan, A., Shyam, P., Sastry, G., Askell, A., et~al.
\newblock Language models are few-shot learners.
\newblock \emph{Advances in neural information processing systems}, 33:\penalty0 1877--1901, 2020.

\bibitem[Buckland \& Gey(1994)Buckland and Gey]{buckland1994relationship}
Buckland, M. and Gey, F.
\newblock The relationship between recall and precision.
\newblock \emph{Journal of the American society for information science}, 45\penalty0 (1):\penalty0 12--19, 1994.

\bibitem[Burges et~al.(2005)Burges, Shaked, Renshaw, Lazier, Deeds, Hamilton, and Hullender]{burges2005learning}
Burges, C., Shaked, T., Renshaw, E., Lazier, A., Deeds, M., Hamilton, N., and Hullender, G.
\newblock Learning to rank using gradient descent.
\newblock In \emph{Proceedings of the 22nd international conference on Machine learning}, pp.\  89--96, 2005.

\bibitem[Cai et~al.(2018)Cai, Wang, Bi, Tu, Liu, Lam, and Shi]{cai2018skeleton}
Cai, D., Wang, Y., Bi, V., Tu, Z., Liu, X., Lam, W., and Shi, S.
\newblock Skeleton-to-response: Dialogue generation guided by retrieval memory.
\newblock \emph{arXiv preprint arXiv:1809.05296}, 2018.

\bibitem[Cai et~al.(2019)Cai, Wang, Bi, Tu, Liu, and Shi]{cai2019retrieval}
Cai, D., Wang, Y., Bi, W., Tu, Z., Liu, X., and Shi, S.
\newblock Retrieval-guided dialogue response generation via a matching-to-generation framework.
\newblock In \emph{Proceedings of the 2019 Conference on Empirical Methods in Natural Language Processing and the 9th International Joint Conference on Natural Language Processing (EMNLP-IJCNLP)}, pp.\  1866--1875, 2019.

\bibitem[Cai et~al.(2021)Cai, Wang, Li, Lam, and Liu]{cai2021neural}
Cai, D., Wang, Y., Li, H., Lam, W., and Liu, L.
\newblock Neural machine translation with monolingual translation memory.
\newblock \emph{arXiv preprint arXiv:2105.11269}, 2021.

\bibitem[Cao et~al.(2007)Cao, Qin, Liu, Tsai, and Li]{cao2007learning}
Cao, Z., Qin, T., Liu, T.-Y., Tsai, M.-F., and Li, H.
\newblock Learning to rank: from pairwise approach to listwise approach.
\newblock In \emph{Proceedings of the 24th international conference on Machine learning}, pp.\  129--136, 2007.

\bibitem[Chen et~al.(2021)Chen, Tworek, Jun, Yuan, Pinto, Kaplan, Edwards, Burda, Joseph, Brockman, et~al.]{chen2021evaluating}
Chen, M., Tworek, J., Jun, H., Yuan, Q., Pinto, H. P. d.~O., Kaplan, J., Edwards, H., Burda, Y., Joseph, N., Brockman, G., et~al.
\newblock Evaluating large language models trained on code.
\newblock \emph{arXiv preprint arXiv:2107.03374}, 2021.

\bibitem[Chowdhery et~al.(2023)Chowdhery, Narang, Devlin, Bosma, Mishra, Roberts, Barham, Chung, Sutton, Gehrmann, et~al.]{chowdhery2023palm}
Chowdhery, A., Narang, S., Devlin, J., Bosma, M., Mishra, G., Roberts, A., Barham, P., Chung, H.~W., Sutton, C., Gehrmann, S., et~al.
\newblock Palm: Scaling language modeling with pathways.
\newblock \emph{Journal of Machine Learning Research}, 24\penalty0 (240):\penalty0 1--113, 2023.

\bibitem[Devlin et~al.(2018)Devlin, Chang, Lee, and Toutanova]{devlin2018bert}
Devlin, J., Chang, M.-W., Lee, K., and Toutanova, K.
\newblock Bert: Pre-training of deep bidirectional transformers for language understanding.
\newblock \emph{arXiv preprint arXiv:1810.04805}, 2018.

\bibitem[Dinan et~al.(2018)Dinan, Roller, Shuster, Fan, Auli, and Weston]{dinan2018wizard}
Dinan, E., Roller, S., Shuster, K., Fan, A., Auli, M., and Weston, J.
\newblock Wizard of wikipedia: Knowledge-powered conversational agents.
\newblock \emph{arXiv preprint arXiv:1811.01241}, 2018.

\bibitem[Dong et~al.(2022)Dong, Ni, Bikel, Alfonseca, Wang, Qu, and Zitouni]{dong2022exploring}
Dong, Z., Ni, J., Bikel, D.~M., Alfonseca, E., Wang, Y., Qu, C., and Zitouni, I.
\newblock Exploring dual encoder architectures for question answering.
\newblock \emph{arXiv preprint arXiv:2204.07120}, 2022.

\bibitem[Freund et~al.(2003)Freund, Iyer, Schapire, and Singer]{freund2003efficient}
Freund, Y., Iyer, R., Schapire, R.~E., and Singer, Y.
\newblock An efficient boosting algorithm for combining preferences.
\newblock \emph{Journal of machine learning research}, 4\penalty0 (Nov):\penalty0 933--969, 2003.

\bibitem[Gao et~al.(2023)Gao, Xiong, Gao, Jia, Pan, Bi, Dai, Sun, and Wang]{gao2023retrieval}
Gao, Y., Xiong, Y., Gao, X., Jia, K., Pan, J., Bi, Y., Dai, Y., Sun, J., and Wang, H.
\newblock Retrieval-augmented generation for large language models: A survey.
\newblock \emph{arXiv preprint arXiv:2312.10997}, 2023.

\bibitem[Gillick et~al.(2018)Gillick, Presta, and Tomar]{gillick2018end}
Gillick, D., Presta, A., and Tomar, G.~S.
\newblock End-to-end retrieval in continuous space.
\newblock \emph{arXiv preprint arXiv:1811.08008}, 2018.

\bibitem[Gu et~al.(2018)Gu, Wang, Cho, and Li]{gu2018search}
Gu, J., Wang, Y., Cho, K., and Li, V.~O.
\newblock Search engine guided neural machine translation.
\newblock In \emph{Proceedings of the AAAI Conference on Artificial Intelligence}, volume~32, 2018.

\bibitem[Guo et~al.(2020)Guo, Yang, Cer, Shen, and Constant]{guo2020multireqa}
Guo, M., Yang, Y., Cer, D., Shen, Q., and Constant, N.
\newblock Multireqa: A cross-domain evaluation for retrieval question answering models.
\newblock \emph{arXiv preprint arXiv:2005.02507}, 2020.

\bibitem[Gupta et~al.(2024)Gupta, Shirgaonkar, Balaguer, Silva, Holstein, Li, Marsman, Nunes, Rouzbahman, Sharp, et~al.]{gupta2024rag}
Gupta, A., Shirgaonkar, A., Balaguer, A. d.~L., Silva, B., Holstein, D., Li, D., Marsman, J., Nunes, L.~O., Rouzbahman, M., Sharp, M., et~al.
\newblock Rag vs fine-tuning: Pipelines, tradeoffs, and a case study on agriculture.
\newblock \emph{arXiv preprint arXiv:2401.08406}, 2024.

\bibitem[Hashimoto et~al.(2018)Hashimoto, Guu, Oren, and Liang]{hashimoto2018retrieve}
Hashimoto, T.~B., Guu, K., Oren, Y., and Liang, P.~S.
\newblock A retrieve-and-edit framework for predicting structured outputs.
\newblock \emph{Advances in Neural Information Processing Systems}, 31, 2018.

\bibitem[Herbrich et~al.(1999)Herbrich, Graepel, and Obermayer]{herbrich1999support}
Herbrich, R., Graepel, T., and Obermayer, K.
\newblock Support vector learning for ordinal regression.
\newblock 1999.

\bibitem[Izacard et~al.(2021)Izacard, Caron, Hosseini, Riedel, Bojanowski, Joulin, and Grave]{izacard2021unsupervised}
Izacard, G., Caron, M., Hosseini, L., Riedel, S., Bojanowski, P., Joulin, A., and Grave, E.
\newblock Unsupervised dense information retrieval with contrastive learning.
\newblock \emph{arXiv preprint arXiv:2112.09118}, 2021.

\bibitem[Jiang et~al.(2023)Jiang, Xu, Gao, Sun, Liu, Dwivedi-Yu, Yang, Callan, and Neubig]{jiang2023active}
Jiang, Z., Xu, F.~F., Gao, L., Sun, Z., Liu, Q., Dwivedi-Yu, J., Yang, Y., Callan, J., and Neubig, G.
\newblock Active retrieval augmented generation.
\newblock \emph{arXiv preprint arXiv:2305.06983}, 2023.

\bibitem[Karpukhin et~al.(2020)Karpukhin, O{\u{g}}uz, Min, Lewis, Wu, Edunov, Chen, and Yih]{karpukhin2020dense}
Karpukhin, V., O{\u{g}}uz, B., Min, S., Lewis, P., Wu, L., Edunov, S., Chen, D., and Yih, W.-t.
\newblock Dense passage retrieval for open-domain question answering.
\newblock \emph{arXiv preprint arXiv:2004.04906}, 2020.

\bibitem[Lee et~al.(2019)Lee, Chang, and Toutanova]{lee2019latent}
Lee, K., Chang, M.-W., and Toutanova, K.
\newblock Latent retrieval for weakly supervised open domain question answering.
\newblock \emph{arXiv preprint arXiv:1906.00300}, 2019.

\bibitem[Lewis et~al.(2020)Lewis, Perez, Piktus, Petroni, Karpukhin, Goyal, K{\"u}ttler, Lewis, Yih, Rockt{\"a}schel, et~al.]{lewis2020retrieval}
Lewis, P., Perez, E., Piktus, A., Petroni, F., Karpukhin, V., Goyal, N., K{\"u}ttler, H., Lewis, M., Yih, W.-t., Rockt{\"a}schel, T., et~al.
\newblock Retrieval-augmented generation for knowledge-intensive nlp tasks.
\newblock \emph{Advances in Neural Information Processing Systems}, 33:\penalty0 9459--9474, 2020.

\bibitem[Lian et~al.(2019)Lian, Xie, Wang, Peng, and Wu]{lian2019learning}
Lian, R., Xie, M., Wang, F., Peng, J., and Wu, H.
\newblock Learning to select knowledge for response generation in dialog systems.
\newblock \emph{arXiv preprint arXiv:1902.04911}, 2019.

\bibitem[Lightman et~al.(2023)Lightman, Kosaraju, Burda, Edwards, Baker, Lee, Leike, Schulman, Sutskever, and Cobbe]{lightman2023let}
Lightman, H., Kosaraju, V., Burda, Y., Edwards, H., Baker, B., Lee, T., Leike, J., Schulman, J., Sutskever, I., and Cobbe, K.
\newblock Let's verify step by step.
\newblock \emph{arXiv preprint arXiv:2305.20050}, 2023.

\bibitem[Liu et~al.(2009)]{liu2009learning}
Liu, T.-Y. et~al.
\newblock Learning to rank for information retrieval.
\newblock \emph{Foundations and Trends{\textregistered} in Information Retrieval}, 3\penalty0 (3):\penalty0 225--331, 2009.

\bibitem[Luan et~al.(2021)Luan, Eisenstein, Toutanova, and Collins]{luan2021sparse}
Luan, Y., Eisenstein, J., Toutanova, K., and Collins, M.
\newblock Sparse, dense, and attentional representations for text retrieval.
\newblock \emph{Transactions of the Association for Computational Linguistics}, 9:\penalty0 329--345, 2021.

\bibitem[Luce(2005)]{luce2005individual}
Luce, R.~D.
\newblock \emph{Individual choice behavior: A theoretical analysis}.
\newblock Courier Corporation, 2005.

\bibitem[Lyu et~al.(2024)Lyu, Li, Niu, Xiong, Tang, Wang, Wu, Liu, Xu, and Chen]{lyu2024crud}
Lyu, Y., Li, Z., Niu, S., Xiong, F., Tang, B., Wang, W., Wu, H., Liu, H., Xu, T., and Chen, E.
\newblock Crud-rag: A comprehensive chinese benchmark for retrieval-augmented generation of large language models.
\newblock \emph{arXiv preprint arXiv:2401.17043}, 2024.

\bibitem[Maia et~al.(2018)Maia, Handschuh, Freitas, Davis, McDermott, Zarrouk, and Balahur]{maia201818}
Maia, M., Handschuh, S., Freitas, A., Davis, B., McDermott, R., Zarrouk, M., and Balahur, A.
\newblock Www'18 open challenge: financial opinion mining and question answering.
\newblock In \emph{Companion proceedings of the the web conference 2018}, pp.\  1941--1942, 2018.

\bibitem[Manakul et~al.(2023)Manakul, Liusie, and Gales]{manakul2023selfcheckgpt}
Manakul, P., Liusie, A., and Gales, M.~J.
\newblock Selfcheckgpt: Zero-resource black-box hallucination detection for generative large language models.
\newblock \emph{arXiv preprint arXiv:2303.08896}, 2023.

\bibitem[Muennighoff et~al.(2022)Muennighoff, Tazi, Magne, and Reimers]{muennighoff2022mteb}
Muennighoff, N., Tazi, N., Magne, L., and Reimers, N.
\newblock Mteb: Massive text embedding benchmark.
\newblock \emph{arXiv preprint arXiv:2210.07316}, 2022.

\bibitem[Oord et~al.(2018)Oord, Li, and Vinyals]{oord2018representation}
Oord, A. v.~d., Li, Y., and Vinyals, O.
\newblock Representation learning with contrastive predictive coding.
\newblock \emph{arXiv preprint arXiv:1807.03748}, 2018.

\bibitem[Ovadia et~al.(2023)Ovadia, Brief, Mishaeli, and Elisha]{ovadia2023fine}
Ovadia, O., Brief, M., Mishaeli, M., and Elisha, O.
\newblock Fine-tuning or retrieval? comparing knowledge injection in llms.
\newblock \emph{arXiv preprint arXiv:2312.05934}, 2023.

\bibitem[Paszke et~al.(2019)Paszke, Gross, Massa, Lerer, Bradbury, Chanan, Killeen, Lin, Gimelshein, Antiga, et~al.]{paszke2019pytorch}
Paszke, A., Gross, S., Massa, F., Lerer, A., Bradbury, J., Chanan, G., Killeen, T., Lin, Z., Gimelshein, N., Antiga, L., et~al.
\newblock Pytorch: An imperative style, high-performance deep learning library.
\newblock \emph{Advances in neural information processing systems}, 32, 2019.

\bibitem[Plackett(1975)]{plackett1975analysis}
Plackett, R.~L.
\newblock The analysis of permutations.
\newblock \emph{Journal of the Royal Statistical Society Series C: Applied Statistics}, 24\penalty0 (2):\penalty0 193--202, 1975.

\bibitem[Qin et~al.(2019)Qin, Galley, Brockett, Liu, Gao, Dolan, Choi, and Gao]{qin2019conversing}
Qin, L., Galley, M., Brockett, C., Liu, X., Gao, X., Dolan, B., Choi, Y., and Gao, J.
\newblock Conversing by reading: Contentful neural conversation with on-demand machine reading.
\newblock \emph{arXiv preprint arXiv:1906.02738}, 2019.

\bibitem[Robertson et~al.(2009)Robertson, Zaragoza, et~al.]{robertson2009probabilistic}
Robertson, S., Zaragoza, H., et~al.
\newblock The probabilistic relevance framework: Bm25 and beyond.
\newblock \emph{Foundations and Trends{\textregistered} in Information Retrieval}, 3\penalty0 (4):\penalty0 333--389, 2009.

\bibitem[Shi et~al.(2023)Shi, Min, Yasunaga, Seo, James, Lewis, Zettlemoyer, and Yih]{shi2023replug}
Shi, W., Min, S., Yasunaga, M., Seo, M., James, R., Lewis, M., Zettlemoyer, L., and Yih, W.-t.
\newblock Replug: Retrieval-augmented black-box language models.
\newblock \emph{arXiv preprint arXiv:2301.12652}, 2023.

\bibitem[Shuster et~al.(2021)Shuster, Poff, Chen, Kiela, and Weston]{shuster2021retrieval}
Shuster, K., Poff, S., Chen, M., Kiela, D., and Weston, J.
\newblock Retrieval augmentation reduces hallucination in conversation.
\newblock \emph{arXiv preprint arXiv:2104.07567}, 2021.

\bibitem[Siriwardhana et~al.(2023)Siriwardhana, Weerasekera, Wen, Kaluarachchi, Rana, and Nanayakkara]{siriwardhana2023improving}
Siriwardhana, S., Weerasekera, R., Wen, E., Kaluarachchi, T., Rana, R., and Nanayakkara, S.
\newblock Improving the domain adaptation of retrieval augmented generation (rag) models for open domain question answering.
\newblock \emph{Transactions of the Association for Computational Linguistics}, 11:\penalty0 1--17, 2023.

\bibitem[Su et~al.(2021)Su, Wang, Cai, Baker, Korhonen, and Collier]{su2021prototype}
Su, Y., Wang, Y., Cai, D., Baker, S., Korhonen, A., and Collier, N.
\newblock Prototype-to-style: Dialogue generation with style-aware editing on retrieval memory.
\newblock \emph{IEEE/ACM Transactions on Audio, Speech, and Language Processing}, 29:\penalty0 2152--2161, 2021.

\bibitem[Team et~al.(2023)Team, Anil, Borgeaud, Wu, Alayrac, Yu, Soricut, Schalkwyk, Dai, Hauth, et~al.]{team2023gemini}
Team, G., Anil, R., Borgeaud, S., Wu, Y., Alayrac, J.-B., Yu, J., Soricut, R., Schalkwyk, J., Dai, A.~M., Hauth, A., et~al.
\newblock Gemini: a family of highly capable multimodal models.
\newblock \emph{arXiv preprint arXiv:2312.11805}, 2023.

\bibitem[Thakur et~al.(2021)Thakur, Reimers, R{\"u}ckl{\'e}, Srivastava, and Gurevych]{thakur2021beir}
Thakur, N., Reimers, N., R{\"u}ckl{\'e}, A., Srivastava, A., and Gurevych, I.
\newblock Beir: A heterogenous benchmark for zero-shot evaluation of information retrieval models.
\newblock \emph{arXiv preprint arXiv:2104.08663}, 2021.

\bibitem[Valizadegan et~al.(2009)Valizadegan, Jin, Zhang, and Mao]{valizadegan2009learning}
Valizadegan, H., Jin, R., Zhang, R., and Mao, J.
\newblock Learning to rank by optimizing ndcg measure.
\newblock \emph{Advances in neural information processing systems}, 22, 2009.

\bibitem[Wu et~al.(2023)Wu, Zhu, Xu, Shum, Niu, Zhong, Song, and Zhang]{wu2023ragtruth}
Wu, Y., Zhu, J., Xu, S., Shum, K., Niu, C., Zhong, R., Song, J., and Zhang, T.
\newblock Ragtruth: A hallucination corpus for developing trustworthy retrieval-augmented language models.
\newblock \emph{arXiv preprint arXiv:2401.00396}, 2023.

\bibitem[Xia et~al.(2008)Xia, Liu, Wang, Zhang, and Li]{xia2008listwise}
Xia, F., Liu, T.-Y., Wang, J., Zhang, W., and Li, H.
\newblock Listwise approach to learning to rank: theory and algorithm.
\newblock In \emph{Proceedings of the 25th international conference on Machine learning}, pp.\  1192--1199, 2008.

\bibitem[Xia et~al.(2019)Xia, Huang, Liu, and Shi]{xia2019graph}
Xia, M., Huang, G., Liu, L., and Shi, S.
\newblock Graph based translation memory for neural machine translation.
\newblock In \emph{Proceedings of the AAAI conference on artificial intelligence}, volume~33, pp.\  7297--7304, 2019.

\bibitem[Xiao et~al.(2023)Xiao, Liu, Zhang, and Muennighof]{xiao2023c}
Xiao, S., Liu, Z., Zhang, P., and Muennighof, N.
\newblock C-pack: Packaged resources to advance general chinese embedding.
\newblock \emph{arXiv preprint arXiv:2309.07597}, 2023.

\end{thebibliography}
\bibliographystyle{bib}

\appendix

\end{document}